\documentclass[10pt,twocolumn,letterpaper]{article}

\usepackage[pagenumbers]{cvpr} % To force page numbers, e.g. for an arXiv version

\pdfoutput=1
\usepackage[dvipsnames]{xcolor}

\definecolor{cvprblue}{rgb}{0.21,0.49,0.74}
\usepackage[pagebackref,breaklinks,colorlinks,citecolor=cvprblue]{hyperref}

\usepackage{array}
\usepackage{algorithm}
\usepackage{algorithmic}

\def\paperID{4426} % *** Enter the Paper ID here
\def\confName{CVPR}
\def\confYear{2024}
\definecolor{ballblue}{rgb}{0.13, 0.67, 0.8}

\title{Visual Programming for Zero-shot Open-Vocabulary 3D Visual Grounding}

\author{Zhihao Yuan$^{1,2}$, Jinke Ren$^{1,2}$, Chun-Mei Feng$^{3}$, Hengshuang Zhao$^{4}$, Shuguang Cui$^{2,1}$, Zhen Li$^{2,1}$\thanks{Corresponding author.} \\
\\
$^{1}$ The Future Network of Intelligence Institute, The Chinese University of Hong Kong (Shenzhen) \\
$^{2}$ School of Science and Engineering, The Chinese University of Hong Kong (Shenzhen) \\
$^{3}$ IHPC, A*STAR, Singapore~~~~~~~~
$^{4}$ The University of Hong Kong \\
{\tt\small \{zhihaoyuan@link.,lizhen@\}cuhk.edu.cn}
}

\begin{document}
\maketitle
\begin{abstract}
3D Visual Grounding (3DVG) aims at localizing 3D object based on textual descriptions. 
Conventional supervised methods for 3DVG often necessitate extensive annotations and a predefined vocabulary, which can be restrictive.
To address this issue, we propose a novel visual programming approach for zero-shot open-vocabulary 3DVG, leveraging the capabilities of large language models (LLMs).
Our approach begins with a unique dialog-based method, engaging with LLMs to establish a foundational understanding of zero-shot 3DVG.
Building on this, we design a visual program that consists of three types of modules, i.e., view-independent, view-dependent, and functional modules.
These modules, specifically tailored for 3D scenarios, work collaboratively to perform complex reasoning and inference.
Furthermore, we develop an innovative language-object correlation module to extend the scope of existing 3D object detectors into open-vocabulary scenarios. 
Extensive experiments demonstrate that our zero-shot approach can outperform some supervised baselines, marking a significant stride towards effective 3DVG.
\end{abstract}    
\section{Introduction}
\label{sec:intro}

\begin{figure}[tbp]
    \centering
    \includegraphics[width=\linewidth]{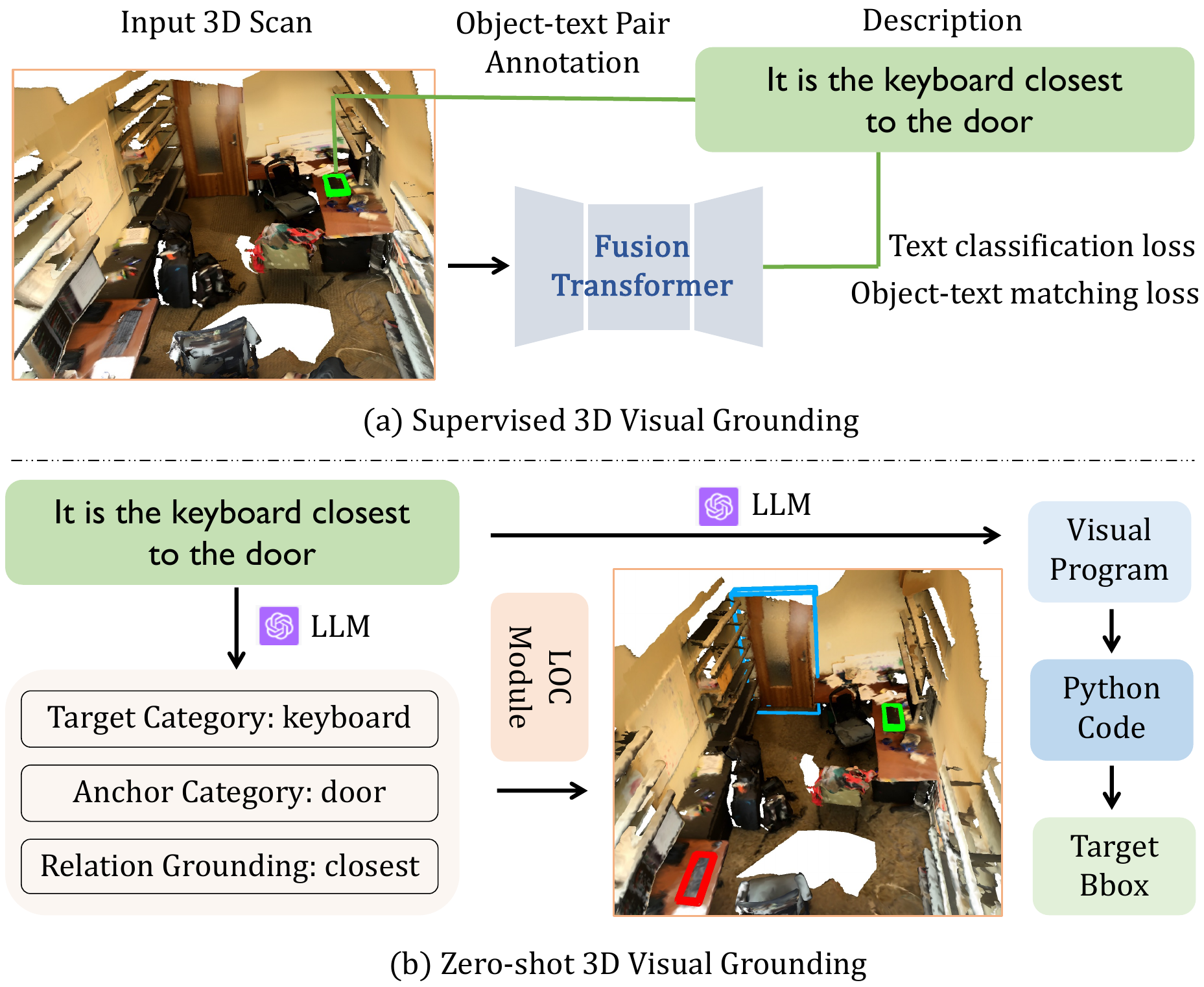}
    \vspace{-15pt}
    \caption{\textbf{Comparative overview  of two 3DVG approaches}, where \textbf{(a)} \textit{Supervised 3DVG} involves input from 3D scans combined with text queries, guided by object-text pair annotations, \textbf{(b)} \textit{Zero-shot 3DVG} identifies the location of target objects using programmatic representation generated by LLMs, \ie, \textit{target category}, \textit{anchor category}, and \textit{relation grounding}, thereby highlighting its superiority in decoding spatial relations and object identifiers within a given space, \eg, the location of the keyboard (outlined in {\color{green}{green}}) can be retrieved based on the distance between the keyboard and the door (outlined in {\color{ballblue}{blue}}).}
    \label{fig:fig1}
    \vspace{-13pt}
\end{figure}

3D Visual Grounding (3DVG) aims to localize specific objects within 3D scenes by using a series of textual descriptions.
This has become a crucial component in a variety of burgeoning applications, such as autonomous robotics \cite{xia2018gibson, wang2019reinforced, feng2021cityflow}, virtual reality \cite{wei2019research, qiu2023trends}, and metaverse \cite{mystakidis2022metaverse, dionisio20133d}. 
For illustration, given a 3D scan in Figure~\ref{fig:fig1}(a) along with its description --- \texttt{It is the keyboard closest to the door}, the goal of 3DVG is to accurately pinpoint the keyboard in the green box, while eliminating potential distractions such as tables and desks. 
Despite the apparent simplicity of this task for humans, it poses a significant challenge for machines due to their inherently limited perceptual capabilities.

Traditional supervised 3DVG approaches \cite{yuan2021instancerefer, yang2021sat, huang2022multi} achieve this objective by leveraging the rich annotations in public datasets, such as ScanRefer \cite{chen2020scanrefer} and Referit3D \cite{achlioptas2020referit3d}.
These approaches typically define 3DVG as a matching problem, generating possible objects via 3D detectors \cite{qi2019deep, jiang2020pointgroup}, and identifying the best match by fusing the visual and textual features.
While these approaches can yield precise results, the acquisition of sufficient annotations is prohibitively resource-intensive for real-world applications. 
Furthermore, these approaches are often constrained by the pre-defined vocabulary during training, making them suboptimal in open-vocabulary scenarios.

To address these issues, we propose a novel visual programming approach for 3DVG that integrates zero-shot learning and large language models (LLMs).
Zero-shot learning \cite{zhang2022pointclip, zhu2023pointclip, zeng2023clip2} can generalize across new categories by leveraging the pre-trained capabilities of CLIP \cite{clip} in the 3D domain.
LLMs \cite{OpenAI_2023, vemprala2023chatgpt, touvron2023llama} can facilitate 3DVG due to their strong planning and reasoning capabilities. 
Regarding this, we first propose a vanilla version {\it{dialog with LLM}}. It describes the location and size of all objects in the scene and instructs the LLM to distinguish the object of interest through an interactive dialog. 
Despite the simplicity of the base approach, the inherent stochasticity and control limitations of LLMs make it hard to capture the {\textbf{view-dependent queries}} and decipher {\textbf{spatial relations in 3D space}}, which are the main challenges of 3DVG. 
To overcome this limitation, we further develop a new visual programming approach, as shown in Figure~\ref{fig:fig1}(b). 
It mainly consists of three steps: (1) generating a 3D visual program using LLMs, (2) interpreting the program into Python code, and (3) identifying the target bounding box by executing the code. 
To enhance the localization accuracy, we further introduce a novel language-object correlation (LOC) module capable of merging the geometric discernment of 3D point clouds with the fine-grained appearance acumen of 2D images. 

In summary, contributions are summarized as follows:
\begin{itemize}
\item We propose an innovative 3D visual programming approach. It eliminates the need for extensive object-text pair annotations required in supervised approaches.

\item We transform the visual program to Python code by designing two types of modules, \ie, relation modules and LOC modules. The former explicitly defines the view-dependent and view-independent relations in 3D space, while the latter captures both the geometric and appearance information for open-vocabulary localization.

\item We conduct extensive experiments on two popular datasets, \ie, ScanRefer~\cite{chen2020scanrefer} and Nr3d~\cite{achlioptas2020referit3d}. We for the first time evaluate the whole validation set rather than a few samples. The results demonstrate the superior performance of our approach, even comparable with existing supervised approaches.
\end{itemize}

%-------------------------------------------------------------------------

\section{Related Work}

\noindent\textbf{Supervised 3DVG.}
3DVG has received much attention in many emerging applications ranging from automatic robotics \cite{xia2018gibson, wang2019reinforced, feng2021cityflow} to metaverse \cite{mystakidis2022metaverse, dionisio20133d}. On the one hand, densely-annotated datasets like ScanRefer \cite{chen2020scanrefer} and Referit3D \cite{achlioptas2020referit3d} can provide well-aligned object-text pairs for ScanNet \cite{scannet}.
On the other hand, most existing methods \cite{yuan2021instancerefer, yang2021sat, roh2021languagerefer, chen2022language, guo2023viewrefer} treat 3DVG as a matching problem, where object identifiers \cite{qi2019deep, jiang2020pointgroup} are utilized to generate candidate objects and find the best-matching one by fusing visual and textual features. Building on this, \cite{wu2023eda, yuan2022toward} attempt to explore the object attributes and relations between different proposals.
Moreover, some works~\cite{chen2023unit3d, jin2023context} have also investigated 3D language pretraining using advanced techniques, such as mask modeling and contrastive learning on paired object-caption data, followed by finetuning on downstream tasks. Additionally, NS3D \cite{hsu2023ns3d} has employed CodeX \cite{chen2021evaluating} to generate hierarchical programs. However, it still needs many data annotations to train the neuro-symbolic networks, thus lacking open-vocabulary and zero-shot capabilities. % 

\begin{figure*}[ht]
    \centering
    \includegraphics[width=\linewidth]{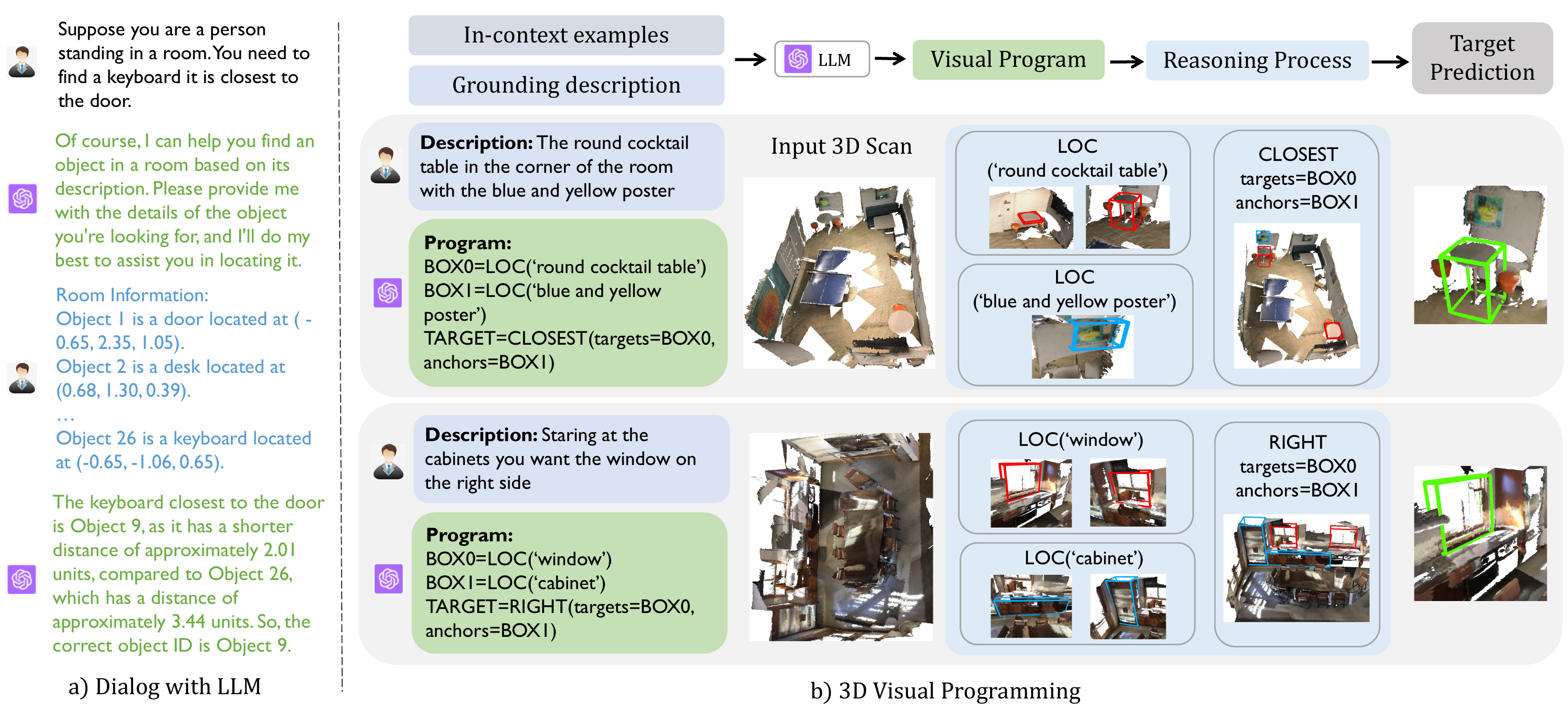}
    \caption{\textbf{Overview of two zero-shot approaches for 3DVG.}  \textbf{(a)} shows the working mechanism of the vanilla dialog with LLM approach. First, we describe the 3DVG task and provide the text descriptions of the room. Then, LLMs identify the objects relevant to the query sentence and perform human-like reasoning. \textbf{(b)} presents the 3D visual programming approach. We first input in-context examples into LLMs. Then, LLMs generate 3D visual programs through the grounding descriptions and perform human-like reasoning. Next, these programs are transformed into executable Python codes via the LOC module for predicting the location of the object. For example, the upper example uses the view-independent module, i.e.,  CLOSEST to determine the proximity in 3D space, while the lower example applies the view-dependent module, i.e.,  RIGHT to establish the relative positioning.}
    \label{fig:fig2}
\end{figure*}

\noindent\textbf{Indoor 3D Scene Understanding.}
3D scene understanding of indoor environments has been widely studied. In specific, the emergence of RBG-D scans datasets \cite{song2015sun, scannet, wald2019rio} greatly push the boundary of several tasks, including 3D object classification \cite{qi2017pointnet, qi2017pointnet++}, 3D object detection \cite{qi2019deep, liu2021group}, 3D semantic segmentation \cite{choy20194d, zhang2020fusion}, 3D instance segmentation \cite{jiang2020pointgroup, vu2022softgroup, schult2023mask3d}, and so on. 
However, these methods are often constrained to a closed set of semantic class labels, limiting their applicability in real-world scenarios.
Recent progress in open-vocabulary image segmentation \cite{ghiasi2022scaling, liang2023open} has inspired research into 3D scene understanding under the open-vocabulary setting. 
For instance, LERF \cite{kerr2023lerf} learns a language field inside NeRF \cite{mildenhall2021nerf} by volume rendering CLIP \cite{clip} features along training rays, enabling it to generate 3D relevancy maps for arbitrary language queries. 
OpenScene \cite{peng2023openscene} extracts image features using 2D open-vocabulary segmentation models \cite{ghiasi2022scaling, li2022languagedriven}, then trained a 3D network to produce point features aligned with multi-view fused pixel features. 
OpenMask3D \cite{takmaz2023openmask3d} utilizes the closed-vocabulary network to generate instance masks while discarded the classification head. 
Despite these advancements, these methods still lack spatial and commonsense reasoning abilities.

\noindent\textbf{LLMs for Vision-Language Tasks.}
Recent progress on LLMs has provided impressive zero-shot planning and reasoning abilities \cite{OpenAI_2023, vemprala2023chatgpt, touvron2023llama}. 
Advanced prompting technologies such as Least-to-Most \cite{zhou2022least}, Think-Step-by-Step \cite{kojima2022large}, and Chain-of-Thought \cite{wei2022chain} are proposed to elicit the capabilities of LLMs. 
These methods can understand human instructions, break complex goals into sub-goals, and control robot agents to execute tasks without additional training \cite{huang2022language, brohan2023can, liang2023code}.
Moreover, when combined with specialized vision models, LLMs can significantly enhance the performance of vision-language tasks. 
For instance, Visual ChatGPT \cite{wu2023visual} uses ChatGPT as a central orchestrator, interfacing with a variety of visual foundation models to solve more challenging problems. 
VISPROG \cite{gupta2023visual} leverages the in-context learning ability to generate high-level modular programs for solving complex and compositional natural language reasoning and image editing tasks. ViperGPT \cite{suris2023vipergpt} directly feeds the API of available modules to LLM and then generates executable Python code for image grounding. 
However, leveraging these capabilities for zero-shot 3D language grounding remains an unexplored area.

\section{Methodology}\label{sec:3}
In Section \ref{sec:3.1}, we introduce the vanilla approach, i.e., dialog with LLM to overcome the annotation issue in 3DVG. From Section \ref{sec:3.2} to Section \ref{sec:3.4}, we present the visual programming approach, address the issue of view-dependent relations, and design the LOC module, respectively.
%-------------------------------------------------------------------------
\subsection{Dialog with LLM}
\label{sec:3.1}
To accomplish the goal of 3DVG, we propose to initiate a dialogue with LLMs. The input for the dialogue consists of a real-world RGB-D scan and a free-form text description $\mathcal{T}$. The text description provides specific information about the target object within a point cloud representation $\mathbf{P} \in \mathbb{R}^{N \times 6}$, where $\mathbf{P}$ is a collection of color-enriched 3D points and $N$ is the total number of such points. The LLM acts as an agent located in the scanned room, which aims to identify the specified object based on the given text description. To bridge the gap between the model's proficiency in understanding text and the spatial nature of the 3DVG task, we first transform the scene into a textual narrative. This narrative can provide a comprehensive account of the objects $\mathcal{O}$ presented in the scene, including their positions and dimensions, which can be expressed as:
\begin{quotation}
\texttt{Object $<$id$>$ is a $<$category$>$ located at (x, y, z) with sizes (width, length, height).}
\end{quotation}

Given such textual layout, we dialog with the LLM by providing the scene's description and query. Our objective is to guide the LLM to identify the object mentioned in the query, while also understand and explain its reasoning process in the identification duration. Particularly, LLM is capable of mimicking the reasoning steps undertaken by humans.  As illustrated in Figure~\ref{fig:fig2}(a), if the LLM get the object information, it can extract the objects relevant to the query sentence, i.e., targets \textit{keyboard} and anchors \textit{door}, and successfully identify the correct target \textit{keyboard} by calculating its distance with \textit{door}.

While LLMs show powerful human-like reasoning capabilties, they still have some limitations. First, it cannot handle the view-dependent issue such as \textit{the right window}. This is becasue the 3D scene can freely rotate to different views while it keeps static in 2D images. LLMs usually make decisions by comparing their x-y values of 3D coordinates despite hinting it in the conversation. Second, mathematical calculation is a common weakness of LLMs but is necessary for 3DVG \cite{dziri2023faith}. For example, in Figure \ref{fig:fig2}(a), distance computing is crucial to solve the \textit{closest} relations, whereas the LLMs cannot always provide accurate results. These two issues stem from LLM's training limitations, which affect the reliability of the dialog with LLM approach.

%-------------------------------------------------------------------------

\subsection{3D Visual Programming}
\label{sec:3.2}
To address the above two issues, we now introduce a new approach that generates visual programs through LLMs. As shown in Figure \ref{fig:fig2}(b), we first construct a set of sample programs to encapsulate human-like problem-solving tactics in 3DVG. Each program includes a sequence of operations, where each operation contains a module name, several input parameters, and an assigned output variable. The output of each step can be reused in the subsequent step, thus creating an interlinked sequence that reflects logical reasoning within a visual context.

We transform the reasoning process of 3DVG into a scripted visual program. Specifically, we collect a set of in-context examples and the corresponding grounding descriptions, and then use LLMs to extrapolate new visual programs tailored to the task. For example, in Figure \ref{fig:fig2}(b), we consider the task prompted by the following description:
\begin{quotation}
\vspace{-0.1cm}
\texttt{The round cocktail table in the corner of the room with the blue and yellow poster.}
\vspace{-0.1cm}
\end{quotation}
In this case, the objective is to identify the round cocktail table, which can be transformed into a operation: BOX0 = $\operatorname{LOC}$(`round cocktail table'), where the LOC operator processes the textual query and outputs the bounding boxes for the target objects. We will elaborate the design of LOC module in Section~\ref{sec:3.4}. Nevertheless, since there may exist multiple similar objects in 3D scenairos, the identified results may not be unique. To overcome this issue, we further pinpoint the blue and yellow poster as an auxiliary reference point by a operation: BOX1 = $\operatorname{LOC}$(`blue and yellow poster'). Then, the CLOSEST module computes the proximity between BOX0 (potential tables) and BOX1 (poster), and selects the table closest to the poster as the result. 

Table \ref{tab:rel} summarizes the common relations in 3DVG. Based on this, we present the detailed visual program by developing three types of modules tailored for 3D contexts:
\begin{itemize}
    \item View-independent modules:   They operate on the 3D spatial relations between objects. For example, the CLOSEST module can discern proximity independent of the viewer's position.
    \item View-dependent modules: They depend on the observer's vantage point. For instance, the RIGHT module determines the right window (TARGET) when looking at cabinets (BOX1) from all windows (BOX0). 
    \item Functional modules: They include multiple operations such as MIN and MAX, which select objects based on the extremal criteria.
\end{itemize}

\begin{table}
    \centering
    \small
    \begin{tabular}{>{\centering\arraybackslash}m{0.3\linewidth}|m{0.6\linewidth}}
    \midrule
        View-independent & \textit{near, close, next to, far, above, below, under, top, on, opposite, middle} \\
    \hline
        View-dependent & \textit{front, behind, back, right, left, facing, leftmost, rightmost, looking, across, between} \\
    \hline
        Functional & \textit{min, max, size, length, width} \\
    \midrule
    \end{tabular}
        \vspace{-8pt}
    \caption{Common relations in 3DVG.}
    \label{tab:rel}
        \vspace{-8pt}
\end{table}

These three types of modules allow the output of one operation to be fed into another operation, thus providing flexible composability. They not only facilitate structured and accurate inference sequences, but also integrate 3D and 2D data to yield a robust and interpretable result for 3DVG.

\begin{figure}
    \centering
    \includegraphics[width=\linewidth]{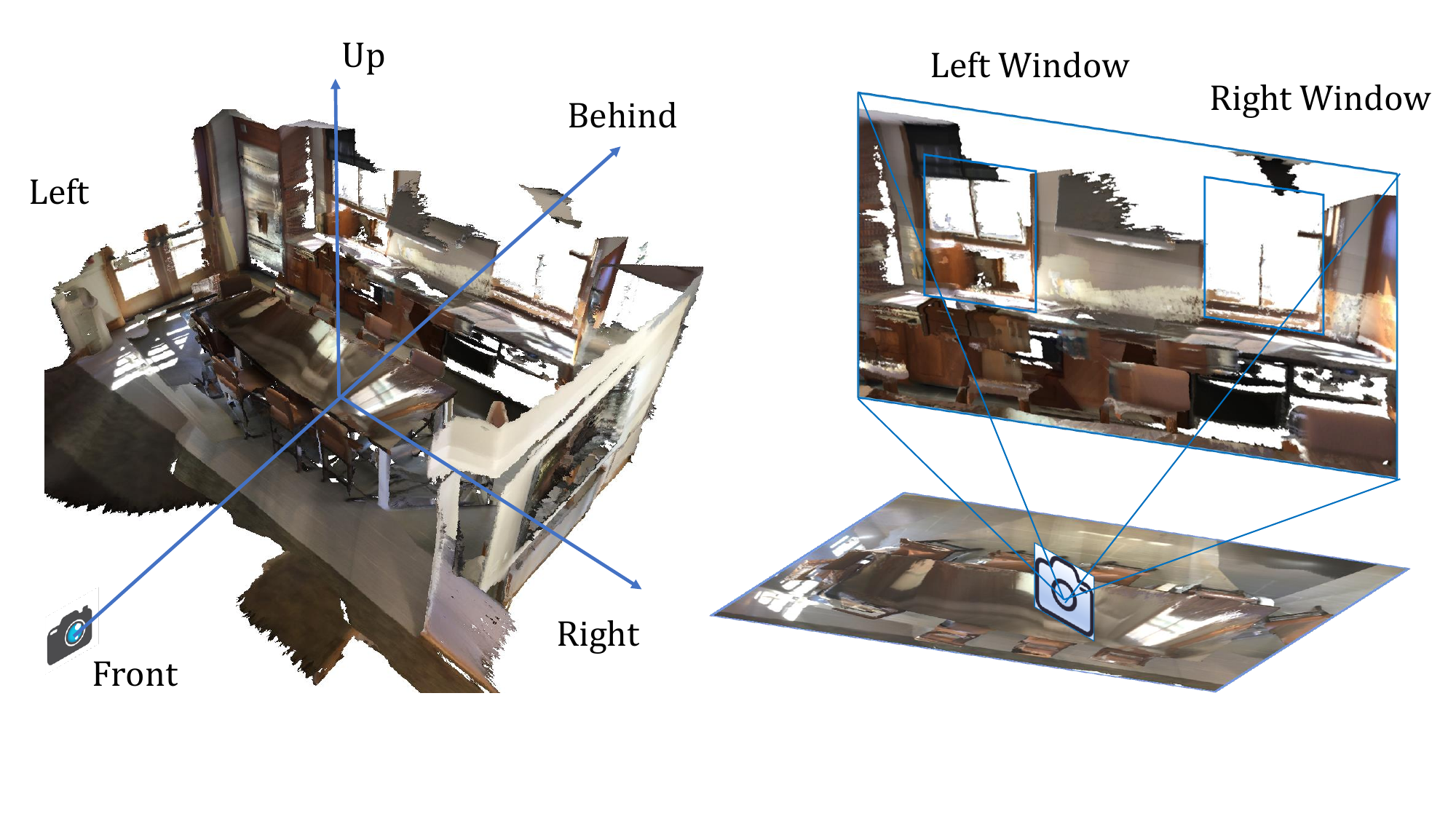}
        \vspace{-25pt}
    \caption{Addressing view-dependent relations: A shift to 2D egocentric view.}
    \label{fig:fig3}
\end{figure}

\subsection{Addressing View-Dependent Relations}
\label{sec:3.3}
In this section, we discuss the intricacies of the view-dependent relations, which are essential for interpreting spatial relations within 3D space. Particularly, the main challenge is the dynamic nature of these relations that will change with the observer's viewpoint. Although traditional supervised approaches can learn these relations implicitly, they cannot provide a definitive resolution.

On 2D planes, the relations, especially the \textit{left} and \textit{right} are well defined. More specifically, \textit{right} often corresponds to the positive direction of the x-axis while the \textit{left} implies the negative direction. Motivated by this, we adopt a 2D egocentric view approach to ensure a consistent frame of reference for the spatial relations in Table \ref{tab:rel}.

Our view-dependent modules  accept a \textit{target} argument and an optional \textit{anchors} parameter. They output the target objects that fulfill the spatial relation to the anchors. When grounding queries do not specify targets, we treat \textit{targets} as \textit{anchors} as well. This approach aligns with our intuition, such as identifying \textit{the left window} by treating all windows themselves as the anchors.

As shown in Figure \ref{fig:fig3}, we assume there is a virtual camera in the center of the room, i.e., $P_{\text{center}}$, which can rotate to align with the location of the anchor objects, i.e.,  $P_{\mathbf{o}_a}$. The 3D objects are projected onto a 2D plane from this vantage point. Assume that the orthogonal camera has a intrinsic matrix $I$, then the 2D projections can be obtained by
\begin{align}
&R, T = \operatorname{Lookat}(P_{\text{center}}, P_{\mathbf{o}_a}, up), \\
&(u, v, w)^{\mathrm{T}} = I\cdot(R|t)\cdot P,
\end{align}
where $\operatorname{Lookat}(\cdot)$ is a view transformation function that computes the rotation matrix $R$ and translation matrix $T$ \cite{vince2006mathematics}, $P=(x,y,z,1)^{\mathrm{T}}$ denotes the 3D coordinate vector, $u$ and $v$ respectively signify the x-axis and y-axis on the 2D plane, and $w$ is the depth value. According to the value of $u$ of an object’s center, we can determine its left or right position --- a lower $u$ value indicates \textit{left}. Similarly, $w$ allows us to distingulish the \textit{front} from \textit{behind}. By synthesizing these concepts, we can define the \textit{between} relation.

The transition from 3D to 2D egocentric perspectiveprovides a clear and consistent solution to interprete view-dependent relations in 3D space, thus enhancing our model’s spatial reasoning ability.

%-------------------------------------------------------------------------

\subsection{Language-Object Correlation  Module}
\label{sec:3.4}
\begin{figure}
    \centering
    \includegraphics[width=\linewidth]{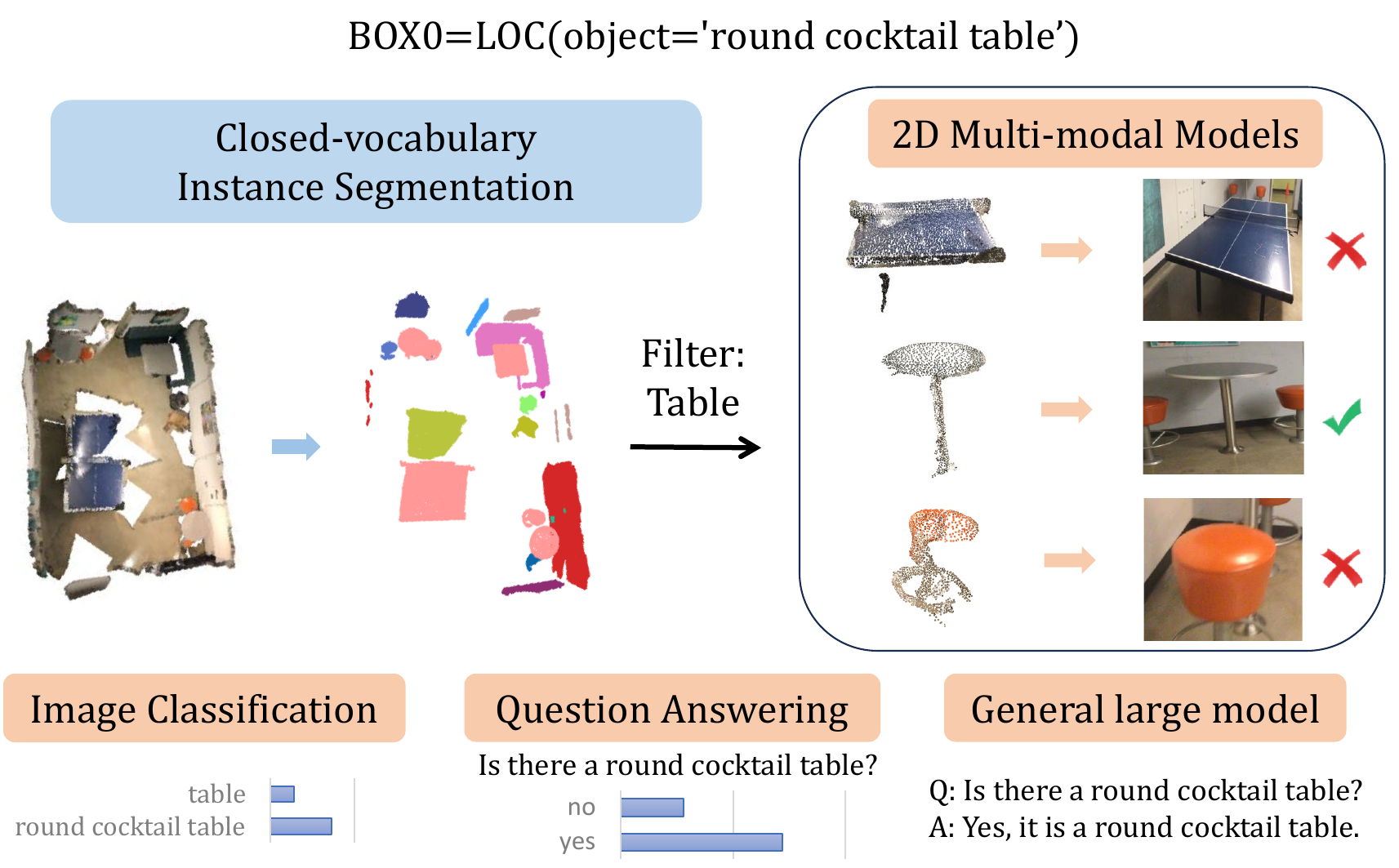}
    \caption{Illustration of the language-object correlation module.}
    \label{fig:fig4}
        \vspace{-12pt}
\end{figure}

Although our zero-shot 3DVG approach does not need extensive grounding annotations, it still requires a basic vision model for object localization. To overcome this issue, previous works \cite{chen2020scanrefer, yuan2021instancerefer}
usually use pre-trained 3D detectors \cite{qi2019deep, jiang2020pointgroup} to generate object proposals and the corresponding labels within a fixed vocabulary. However, this approach is restricted to a predefined object class set, thus limiting the scope of class prediction.
To enable open-vocabulary segmentation, we develop an LOC module, combining the advantages of 3D and 2D networks to extend the labeling capability beyond the closed set. For example, in Figure \ref{fig:fig4}, considering the operation: BOX0 = $\operatorname{LOC}$(`round cocktail table'), we first filter a subset of objects whose predicted label is \textit{table} using a 3D instance segmentation network \cite{schult2023mask3d}. Then we only need to identify a round cocktail table from this subset using the corresponding 2D imagery. By mapping each 3D proposal to its 2D image, we can extract the color and texture details pertinent to our query. To further pinpoint the round cocktail table shown in the Figure \ref{fig:fig4}, we consider three types of 2D multi-modal models:
\begin{itemize}
\item Image classification models: We construct a dynamic vocabulary, including both the query term ``round cocktail table" and the  class ``table" using popular tools such as CLIP \cite{clip}. Then we evaluate the cosine similarity between these terms and the imagery to find the best correlation to our query.
\item Visual question answering models: We raise the question: \texttt{Is there a [query]?} to  the model such as ViLT \cite{kim2021vilt}. Then the model sifts through its dictionary to suggest the most likely answer, i.e., yes or no.
\item General large models: We submit the same inquiry and anticipate a response based on the generated text. This process is crucial for verifying the alignment between the detected table and the query.
\end{itemize}

We shall note that our approach is not limited to specific 3D or 2D models, allowing versatile incorporation of various models. In the experiments, we will demonstrate that the benefit of the LOC modules by comparing with the 3D-only and 2D-only couterparts. Our design indicates a leap forward in 3D open-vocabulary instance segmentation and can improve the object recognition accuracy in 3DVG.

\begin{table*}[h]
    \centering
    \fontsize{9}{9}\selectfont
    \renewcommand{\arraystretch}{1.3}
     \setlength{\tabcolsep}{8pt}
    %\small
    \begin{tabular}{lccccccc}
        \toprule 
        & & \multicolumn{2}{c}{ Unique } & \multicolumn{2}{c}{ Multiple } & \multicolumn{2}{c}{ Overall } \\
         Methods & Supervision & Acc@0.25 & Acc@0.5 & Acc@0.25 & Acc@0.5 & Acc@0.25 & Acc@0.5 \\
        \hline ScanRefer \cite{chen2020scanrefer} & fully & 65.0 & 43.3 & 30.6 & 19.8 & 37.3 & 24.3 \\
        TGNN \cite{huang2021text} & fully & 64.5 & 53.0 & 27.0 & 21.9 & 34.3 & 29.7 \\
        InstanceRefer \cite{yuan2021instancerefer} & fully & 77.5 & 66.8 & 31.3 & 24.8 & 40.2 & 32.9 \\
         3DVG-Transformer \cite{zhao20213dvg} & fully & 81.9 & 60.6 & 39.3 & 28.4 & 47.6 & 34.7 \\
         BUTD-DETR \cite{jain2022bottom} & fully & 84.2 & 66.3 & 46.6 &35.1 & 52.2 & 39.8\\
         \hline
         \hline
         LERF \cite{kerr2023lerf} & - & - & - & -  & - & 4.8 & 0.9 \\
         OpenScene \cite{peng2023openscene} & -  & 20.1 & 13.1 & 11.1 & 4.4 & 13.2 & 6.5 \\
        Ours (2D only) & - & 32.5 & 27.8 & 16.1 & 14.6 & 20.0 & 17.6 \\
        Ours (3D only) & - & 57.1 & 49.4 & 25.9 & 23.3 & 33.1 & 29.3 \\
         Ours & - & \textbf{63.8} & \textbf{58.4} & \textbf{27.7} & \textbf{24.6} & \textbf{36.4} & \textbf{32.7} \\
        \bottomrule
    \end{tabular}
        \vspace{-4pt}
    \caption{3DVG results on ScanRefer validation set. The accuracy on the ``unique" subset, ``multiple" subset, and whole validation set are all provided. Following \cite{chen2020scanrefer}, we label the scene as ``unique" if it only contains a single object of its class. Otherwise, we label it as ``multiple".}
    \label{tab:tab1}
        \vspace{-8pt}
\end{table*}

\begin{table}[h]
    \centering
    % \footnotesize
    \resizebox{\linewidth}{!}{
    \begin{tabular}{lccccc}
        \toprule
        Method & Easy & Hard & Dep. & Indep. & Overall \\
        \hline ReferIt3DNet \cite{achlioptas2020referit3d} & 43.6 & 27.9 & 32.5 & 37.1 & 35.6 \\
        InstanceRefer \cite{yuan2021instancerefer} & 46.0 & 31.8 & 34.5 & 41.9 & 38.8 \\
        3DVG-Transformer \cite{zhao20213dvg} & 48.5 & 34.8 & 34.8 & 43.7 & 40.8 \\
        BUTD-DETR \cite{jain2022bottom} & 60.7 & 48.4 & 46.0 & 58.0 & 54.6 \\
        \hline
        Ours (2D only) & 29.4 & 18.4 & 23.0 & 23.9 & 23.6 \\
        Ours (3D only) & 45.9 & 27.9 & 34.9 & 38.4 & 36.7 \\
        \textbf{Ours} & \textbf{46.5} & \textbf{31.7} & \textbf{36.8} & \textbf{40.0} & \textbf{39.0} \\
        \bottomrule
    \end{tabular}
    }
        \vspace{-4pt}
    \caption{Performance analysis of language grounding on Nr3D. We evaluate the top-1 accuracy using ground-truth boxes.}
    \label{tab:tab2}
        \vspace{-12pt}
\end{table}

\section{Experiments}
\label{sec:exp}

%-------------------------------------------------------------------------
\subsection{Experimental Settings}
\label{sec:4.1}
\textbf{Datasets.}
We use two popular datasets, i.e., ScanRefer \cite{chen2020scanrefer} and  Nr3D \cite{achlioptas2020referit3d} for experiments. ScanRefer is tailored for 3DVG that contains 51,500 sentence descriptions for 800 ScanNet scenes \cite{scannet}. Nr3D is a human-written and free-form dataset for 3DVG, collected by 2-player reference game in 3D scenes. The sentences are divided into ``easy" and ``hard" subsets, where the target object only contains one same-class distractor in the ``easy" subset but contains multiple ones in the ``hard" subset. Depending on whether the sentence requires a specific viewpoint to ground the referred object, the dataset can also be partitioned into ``view depedent" and ``view independent" subsets. For both datasets, we evaluate the zero-shot approaches on the validation split.

\noindent\textbf{Evaluation metrics.}
We consider two settings for performance evaluation. The first one mandates the generation of object proposals, aligning closely with real-world applications. The evaluation metrics are Acc@0.25 and Acc@0.5, representing the percentage of correctly predicted bounding boxes whose IoU exceeds 0.25 or 0.5 with the ground-truth, respectively. This is the default setting for ScanRefer dataset. The second one furnishes ground-truth object masks, necessitating only classification, with an objective to eradicate localization error and achieve high grounding accuracy. This is the default setting for Nr3D dataset.

\noindent\textbf{Baselines.}
We use six supervised and two open-vocabulary 3D scene understanding approaches for performance comparison. For supervised approaches, ScanRefer \cite{chen2020scanrefer} and ReferIt3DNet \cite{achlioptas2020referit3d} encode the 3D point clouds and language separately, and then fuse them to rank the objects by predicted scores. TGNN~\cite{huang2021text} and InstanceRefer~\cite{yuan2021instancerefer} make one further step by learning instance-wise features. 3DVG-Transformer \cite{zhao20213dvg} and BUTD-DETR \cite{jain2022bottom} respectively utilize the Transformer \cite{vaswani2017attention} and DETR \cite{carion2020end} architectures, representing the SoTA approaches.
For open-vocabulary approaches,  OpenScene \cite{peng2023openscene} and LERF \cite{kerr2023lerf} aims to learn a 3D representation aligned with the 2D CLIP feature, thus enabling free-form language grounding. The query $\mathcal{T}$ is processed by the CLIP text encoder, and its similarity is computed against the extracted point features. Finally, they cluster the points with the highest score to determine the target object.

%-------------------------------------------------------------------------

\begin{figure*}[t]
    \centering
    \includegraphics[width=\linewidth]{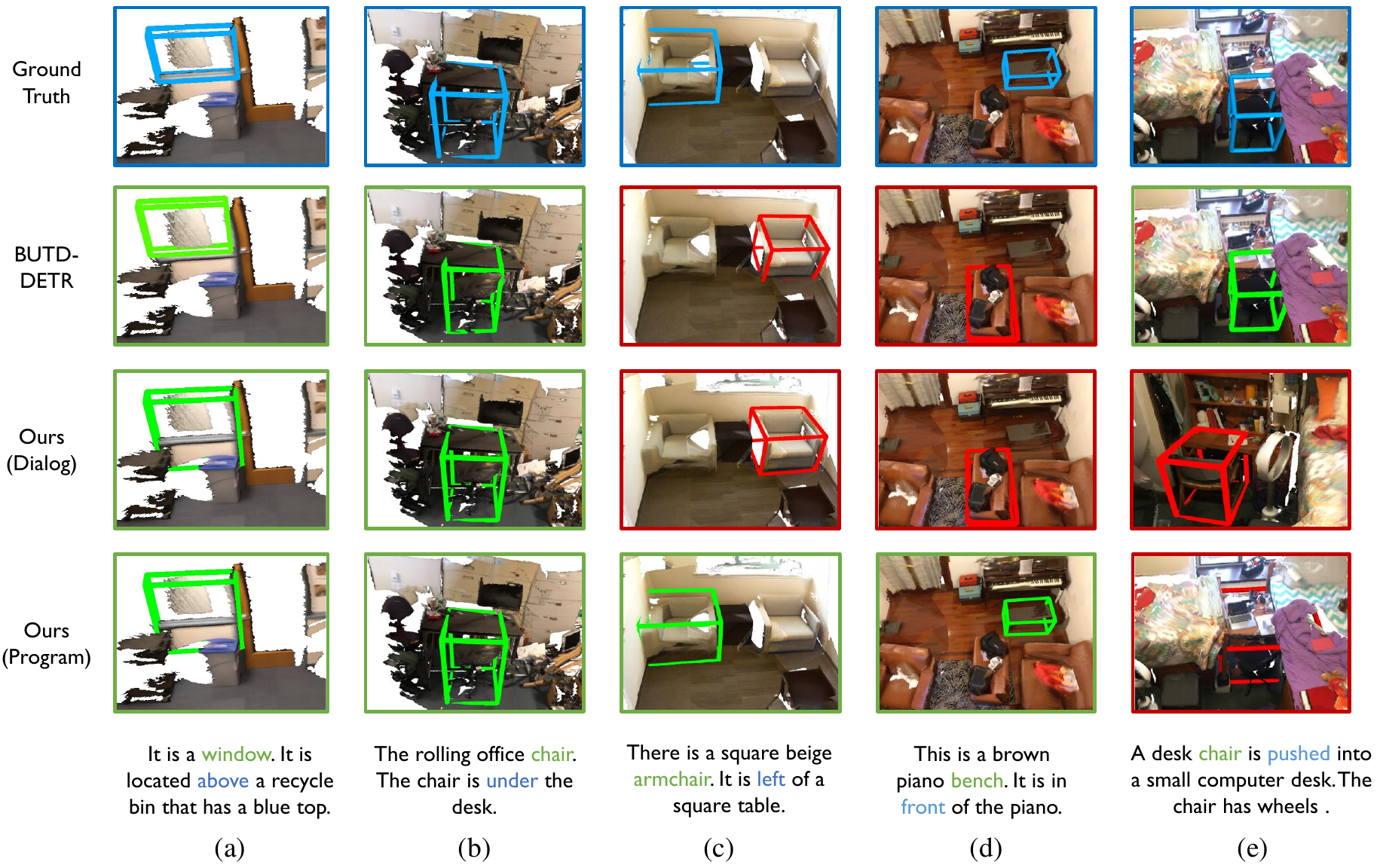}
    \caption{\textbf{Visualization results of 3D visual grounding.} Rendered images of 3D scans are presented, including the ground-truth ({\color{blue}{blue}}), incorrectly identified objects ({\color{red}{red}}), and correctly identified objects ({\color{green}{green}}). 
    }
    \vspace{-12pt}
    \label{fig:vis}
\end{figure*}

\subsection{Quantitative Results}

\textbf{ScanRefer.}
Table \ref{tab:tab1} provides a quantitative assessment of the proposed approach on the ScanRefer  
dataset. We can see that our zero-shot approach outperforms all baseline approaches. Specifically, our approach can achieve a 32.7 Acc@0.5 score, which surpasses the supervised approaches, including the ScanRefer and TGNN. On the other hand, the open-vocabulary approaches LERF and OpenScene can respectively achieve the overall accuracy of 4.8 and 13.2, even with the 0.25 IoU threshold. This is due to their limitations in reasoning and localization precision. Moreover, our zero-shot approach outperforms the approaches that only utilize the 3D or 2D information in the LOC module.
This result demonstrates the effectiveness of incorporating visual programming and perception modules, highlighting our zero-shot approach in navigating the realm of 3DVG.

\noindent\textbf{Nr3D.}
Table \ref{tab:tab2} shows the performance of different approaches on the Nr3D dataset, in which the ground-truth instance mask is also provided. We can see that our zero-shot approach further excels the supervised approach InstanceRefer.
Specifically, our zero-shot approach on the ``view-dependent" split can achieve a 2\% accuracy gain than the 3DVG-Transformer approach.
This performance gain comes from the relation modules, strengthing the potential of our zero-shot approach for 3DVG tasks.

\subsection{Qualitative Results}
Figure \ref{fig:vis} shows the visualizations of the selected samples from the ScanRefer validation set. The four columns present the ground-truth result, the supervised approach BUTD-DETR, the dialog with LLM, and the visual programming approaches, respectively. 
From Figure \ref{fig:vis}(a) and Figure \ref{fig:vis}(b), we can observe that the dialog with LLM and the visual programming approaches can achieve accurate prediction results for view-independent relations, i.e., (\textit{above}, \textit{under}) without much training. On the contrary, both the BUTD-DETR and the dialog with LLM approaches cannot address the view-dependent relations, i.e., (\textit{left}, \textit{front}), as shown in Figure \ref{fig:vis}(c) and Figure \ref{fig:vis}(d). The inherent uncertainty of these relations reflects the limitations of existing methods. However, our visual programming approach can leverage the 2D egocentric views, thus achieving accurate predictions in 3D scenarios.

Figure \ref{fig:vis}(e) presents a failure case, where the dialog with LLM approach cannot recognize \textit{chair has wheels} since it lacks open-vocabulary detection ability. Besides, the visual programming approach makes wrong predictions because the LLM cannot correctly recognize the relation \textit{pushed}. Fortunately, when we correct the program using the CLOSEST module, the visual programming approach can make correct predictions.

\subsection{Ablation Studies}
\label{sec:4.4}
\noindent\textbf{Dialog with LLM vs. visual programming.} We compare the performance of the two proposed zero-shot 3DVG approaches on the ScanRefer validation set with 700 examples. For both approaches, we use two GPT versions, i.e., GPT-3.5-turbo-0613 and GPT-4-0613.
The cost of each GPT version depends on the number of input and output tokens. The experimental results are shown in Table \ref{tab:ablation1}. We can observe that for both zero-shot approaches, GPT4-based approach can achieve higher accuracy than GPT3.5-based approach, even it induces a larger economic cost. On the other hand, the visual programming approach always outperforms the dialog with LLM approach in terms of accuracy and cost, which demonstrates the effectiveness of the proposed visual programming approach. For other experiments, we use GPT3.5 to save cost.

\noindent\textbf{Relation modules.}
We now ablate different relation modules in Section \ref{sec:3.2} to analyze their impact on the system performance. The most important view-dependent and view-independent modules are presented in Table~\ref{tab:5} and \ref{tab:6}, respectively. We can see that LEFT and RIGHT are the most important view-dependent relations, while CLOSEST is the most important view-independent relation. This result is coherent with our motivation and design.
 
\noindent\textbf{LOC module.} We juxtapose our approach by separately omitting the 3D component and 2D component. Both models utilize the instance mask prediction of Mask3D \cite{schult2023mask3d}. Particularly, the 2D-only model solely employs the paired 2D images for classification, while the 3D-only model just uses the 3D result. 
As can be seen from Tables \ref{tab:tab1} and \ref{tab:tab2}, the 2D-only model performs worst when the images of indoor scenes are complicated and have domain gaps with the training samples. 
The 3D-only model performs better since it can utilize the geometric information and is trained on closed-set labels. Our full model can always achieve the best performance because it integrates the geometric distinctiveness of point clouds and the open-vocabulary ability of the image models.
\begin{table}
    \centering
    \small
    \begin{tabular}{lcccc}
        \toprule
        Method & LLM & Acc@0.5 & Tokens & Cost \\
        \hline
        Dialog & GPT3.5 & 25.4 & 1959k & \$3.05 \\
        Dialog & GPT4 & 27.5 & 1916k & \$62.6 \\
        Program & GPT3.5 & 32.1 & 121k & \$0.19 \\
        Program & GPT4 & 35.4 & 115k & \$4.24 \\
        \bottomrule
    \end{tabular}
    \caption{Performance comparison of the dialog with LLM and the visual programming approaches.}
        \vspace{-12pt}
    \label{tab:ablation1}
\end{table}
\begin{table}[t]
    \centering
    \footnotesize
        \setlength\tabcolsep{4.0pt}
        \renewcommand{\arraystretch}{1.2} %
        \begin{tabular}{ccccc|c}
            \hline
             LEFT & RIGHT & FRONT & BEHIND & BETWEEN & Accuracy \\
            \hline
             &  &  &  & & 26.5   \\
            \checkmark &  &  & & & 32.4   \\
             \checkmark & \checkmark &  &  & & 35.9   \\
             \checkmark &  \checkmark & \checkmark & & &  36.8 \\
             \checkmark & \checkmark & \checkmark & \checkmark & &  38.4    \\
              \checkmark & \checkmark & \checkmark & \checkmark & \checkmark & \textbf{39.0}    \\
            \midrule
        \end{tabular}
    \vspace{-8pt}
    \caption{Ablation study of different \textbf{view-dependent} modules.}
    \label{tab:5}%
\end{table}

\begin{table}[t]
    \centering
    \footnotesize
    \renewcommand{\arraystretch}{1.2} %
        \begin{tabular}{cccc|c}
        \hline
             CLOSEST & FARTHEST & LOWER & HIGHER & Accuracy \\
            \hline
             &  &  &  & 18.8   \\
            \checkmark &  &  &  & 30.7   \\
             \checkmark & \checkmark &  &  & 34.0   \\
             \checkmark &  \checkmark & \checkmark & & 36.8 \\
             \checkmark & \checkmark & \checkmark & \checkmark & \textbf{39.0}    \\
            \midrule
        \end{tabular}
    \vspace{-8pt}
    \caption{Ablation study of different \textbf{view-independent} modules.}
    \label{tab:6}%
    \vspace{-12pt}
\end{table}

\noindent\textbf{Generalization.}
As discussed in Section~\ref{sec:3.4}, our framework has strong adaptability for a spectrum of 3D and 2D perception models. To validate this claim, we conduct experiments using several representative models. For 3D perception, we utilize three backbones, i.e., PointNet++ \cite{qi2017pointnet++}, PointNeXt \cite{qian2022pointnext}, and PointBERT \cite{yu2022pointbert}. For 2D perception, we use an image classification model proposed in \cite{clip}, a visual question answering model in \cite{kim2021vilt}, and a general large model BLIP-2 \cite{li2023blip} for testing.
The results are shown in Tables~\ref{table:2d} and~\ref{table:3d}. We can observe that our framework is compatible with other models. Also, it can leverage the advancements within both 2D and 3D foundational models to improve the performance. This cross-model effectiveness demonstrates the robustness and future-proof nature of our approach in the ever-evolving landscape of visual perception models.

\begin{table}[h]
    \centering
     \small
    \begin{tabular}{c|ccc}
        \hline
        2D Assistance & Unique &  Multiple  & Acc@0.25\\
        \hline 
        CLIP & 62.5 & 27.1 & 35.7 \\
        ViLT & 60.3 & 27.1 & 35.1 \\
        BLIP-2 & 63.8 & 27.7 & 36.4 \\
    \midrule
    \end{tabular}
        \vspace{-6pt}
    \caption{Ablation study on different 2D models.}
    \label{table:2d}
        \vspace{-8pt}
\end{table}

    \vspace{-8pt}
\begin{table}[h]
    \centering
     \small
    \begin{tabular}{c|ccc}
     \hline
        3D Backbone & View-dep. &  View-indep. & Overall \\
        \hline 
        PointNet++ & 35.8 & 39.4 & 38.2 \\
        PointBert & 36.0 & 39.8 & 38.6 \\
        PointNeXt & 36.8 & 40.0 & 39.0 \\
    \midrule
    \end{tabular}
    \vspace{-8pt}
    \caption{Ablation study on  different 3D backbones.}
    \label{table:3d}
\end{table}

\begin{figure}
    \centering
    \includegraphics[width=\linewidth]{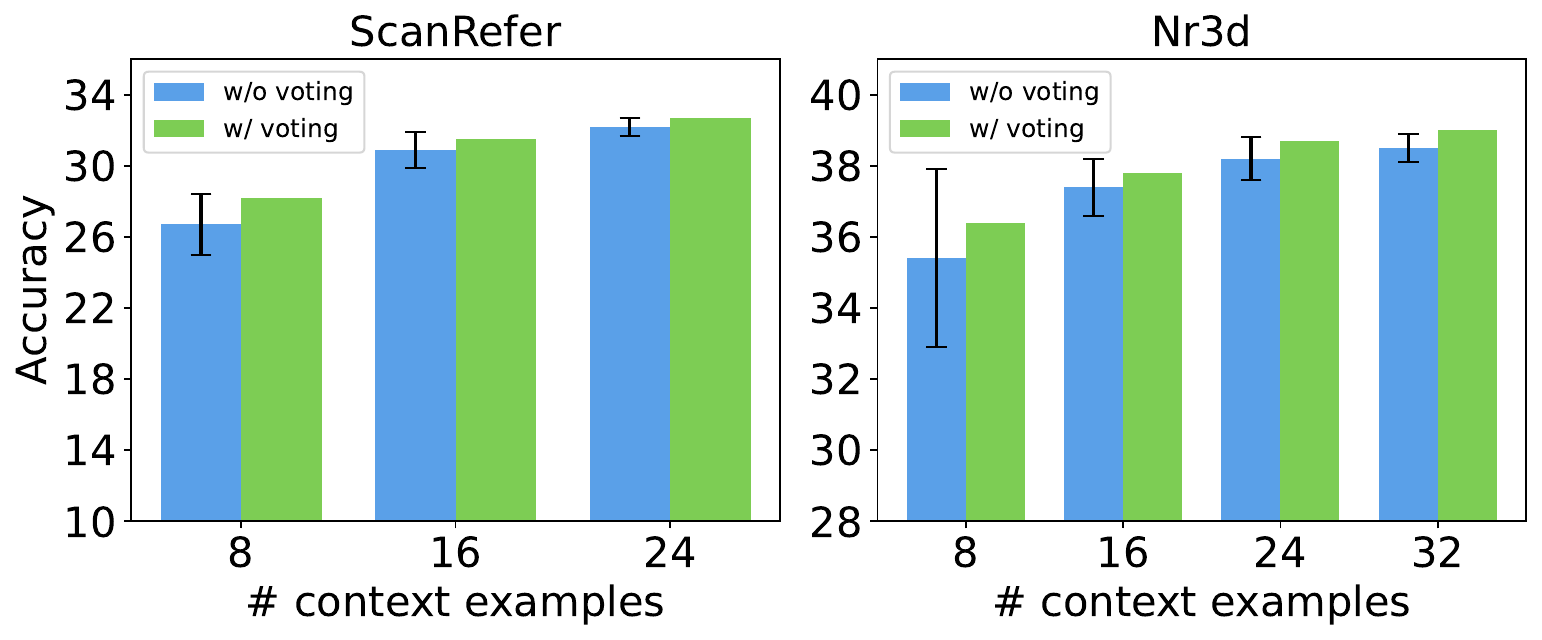}
    \vspace{-15pt}
    \caption{Ablation study on the number of in-context examples. 
    The performance on Nr3D and ScanRefer improves with the number of in-context examples.
    }
    \label{fig:fig5}
    %\vspace{-10pt}
\end{figure}

    \vspace{-8pt}
\noindent\textbf{Effect of prompt size.}
We use different numbers of in-context examples in the prompt for program generation. The result is shown in Figure \ref{fig:fig5}. It can be seen that the performance on ScanRefer and Nr3d improve with the number of examples. This is because more examples can guide LLMs to handle more cases in the visual program generation process. Meanwhile, it still follows the law of diminishing marginal utility. Moreover, we test the voting technique \cite{gupta2023visual} to aggregate the results from multiple runs, which brings some performance gains.

\begin{figure}
    \centering
    \includegraphics[width=\linewidth]{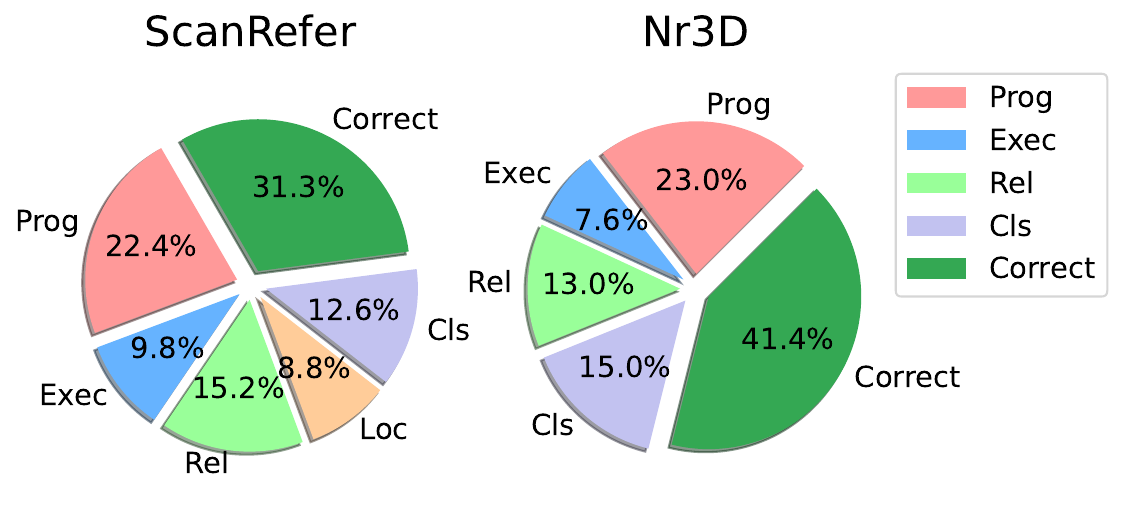}
    %\vspace{-4pt}
    \caption{Breakdown of error sources.}
    \label{fig:error}
    %\vspace{-14pt}
\end{figure}

\noindent\textbf{Error analysis.}
To better understand the limitations of our framework, we conduct error analysis in the following. For each dataset, we select a representative subset with around 100 samples and manually check the rationales offered by the visual program. This introspective method helps identify the dominant error sources and provide guidance to improve our framework. The result is illustrated in Figure \ref{fig:error}, which reveals that the generation of accurate visual programs is the primary error source. Therefore, the performance can be improved by using more in-context examples and more powerful LLMs. 
The second error source is the object localization and classification, indicating that object detection and classification in 3D space remains a critical component. Additionally, the results point out a need for developing additional modules to handle a wider array of spatial relations like ``opposite". These issues have not been well addressed in current framework.

\section{Conclusion}
In this paper, we present a novel zero-shot approach for 3DVG to eliminate the need for extensive annotations and predefined vocabularies. A vanilla dialog with LLM approach is first proposed by taking interactive dialog with LLMs. A visual programming approach is further developed, which leverages three types of modules to navigate the intricate 3D relations. To adapt to open-vocabulary scenarios, we also develop a LOC module to seamlessly integrate both 3D and 2D features. Experimental results demonstrate the superiority of the proposed approach and highlight its potential to advance the field of 3DVG.

{
    \small
    \bibliographystyle{ieeenat_fullname}
    \bibliography{main}
}

 \clearpage
\setcounter{page}{1}
\maketitlesupplementary

\section*{Contents}
The following two items are included in the supplementary material:
\begin{itemize}
  \item Visualization examples for zero-shot 3DVG in Section~\ref{sec:1}.
  \item Prompts for generating 3D visual program in Section~\ref{sec:2}. 
    \vspace{10pt}
\end{itemize}
\begin{figure}[h]
    \centering
\includegraphics[width=\linewidth]{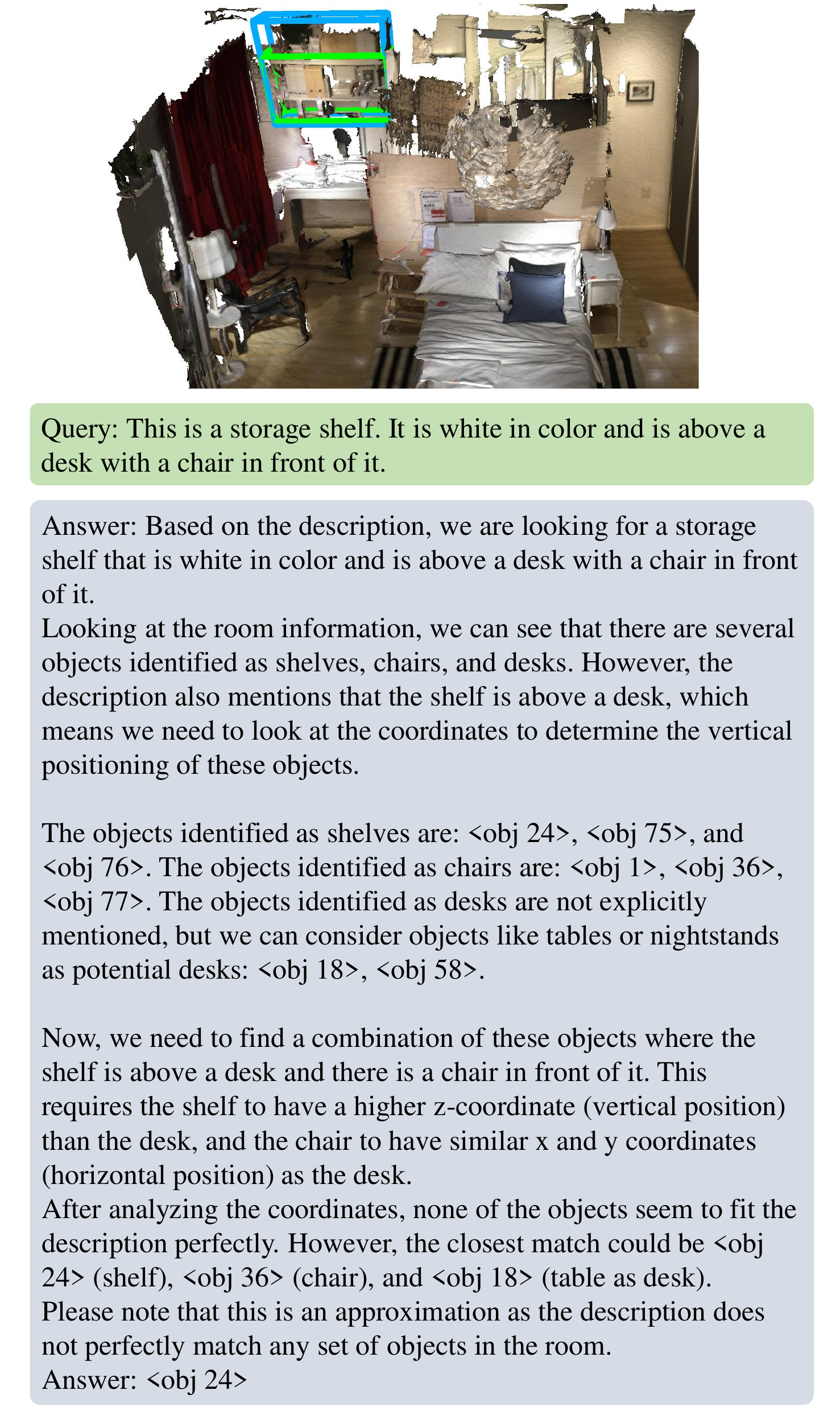}
        \vspace{-20pt}
    \caption{Visualization example (a) for zero-shot 3DVG.}
    \label{fig:supp_1}
\end{figure}
\begin{figure}[h]
    \centering
    \includegraphics[width=\linewidth]{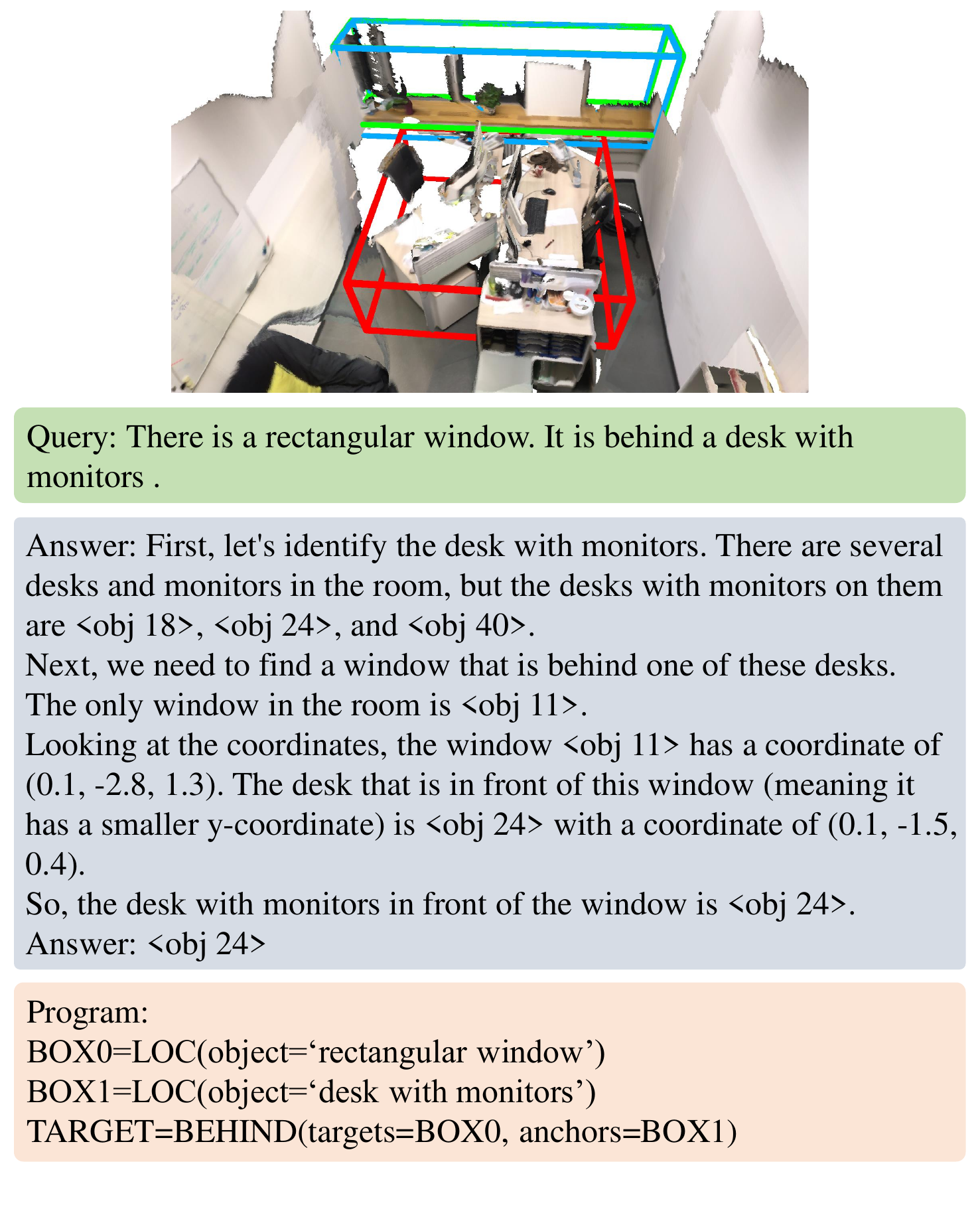}
        \vspace{-20pt}
    \caption{Visualization example (b) for zero-shot 3DVG.}
    \label{fig:supp_2}
\end{figure}
\begin{figure}[h]
    \centering
    \includegraphics[width=\linewidth]{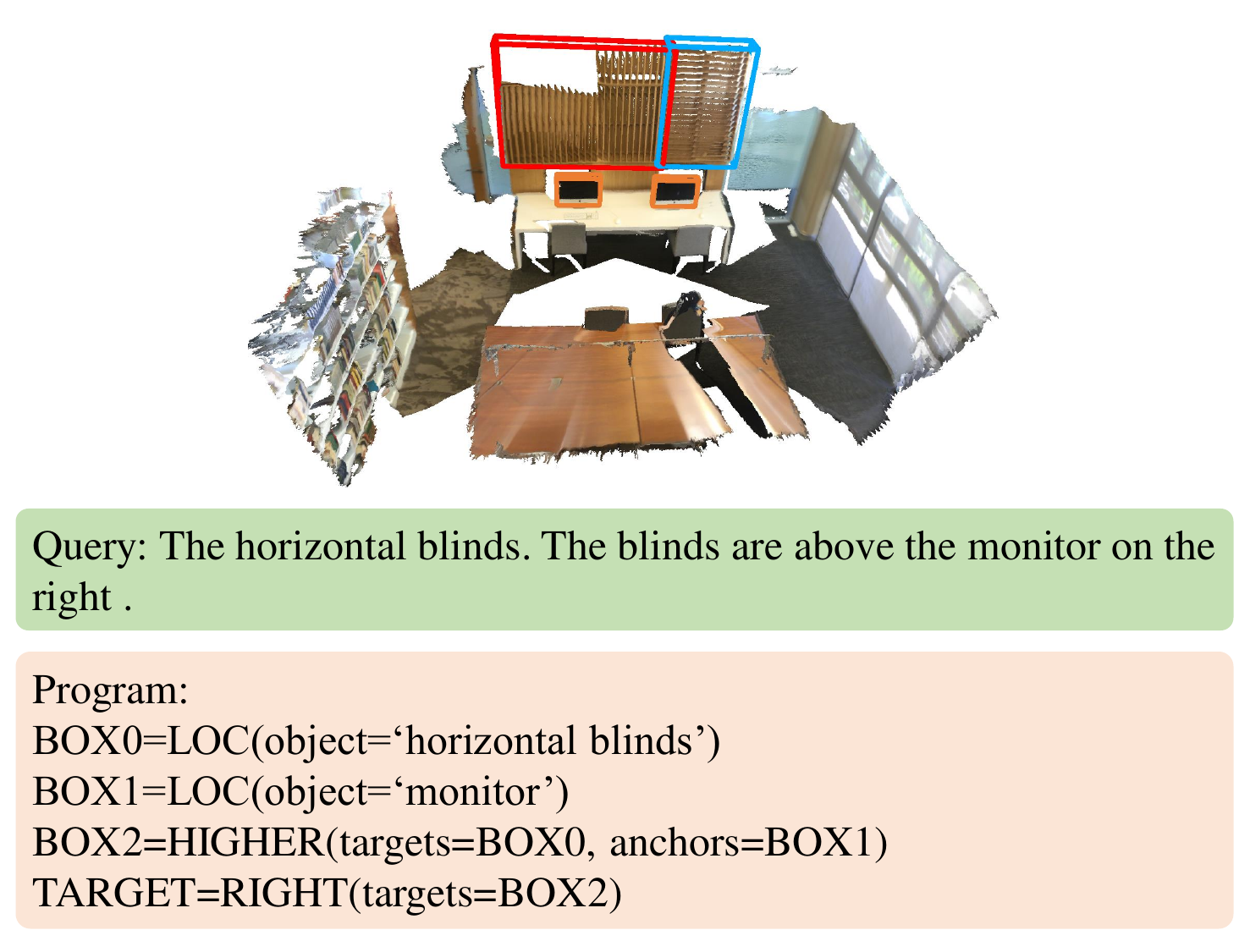}
        \vspace{-20pt}
    \caption{Visualization example (c) for zero-shot 3DVG.}
    \label{fig:supp_3}
\end{figure}

\begin{figure*}[t]
    \centering
\includegraphics[width=5.8in]{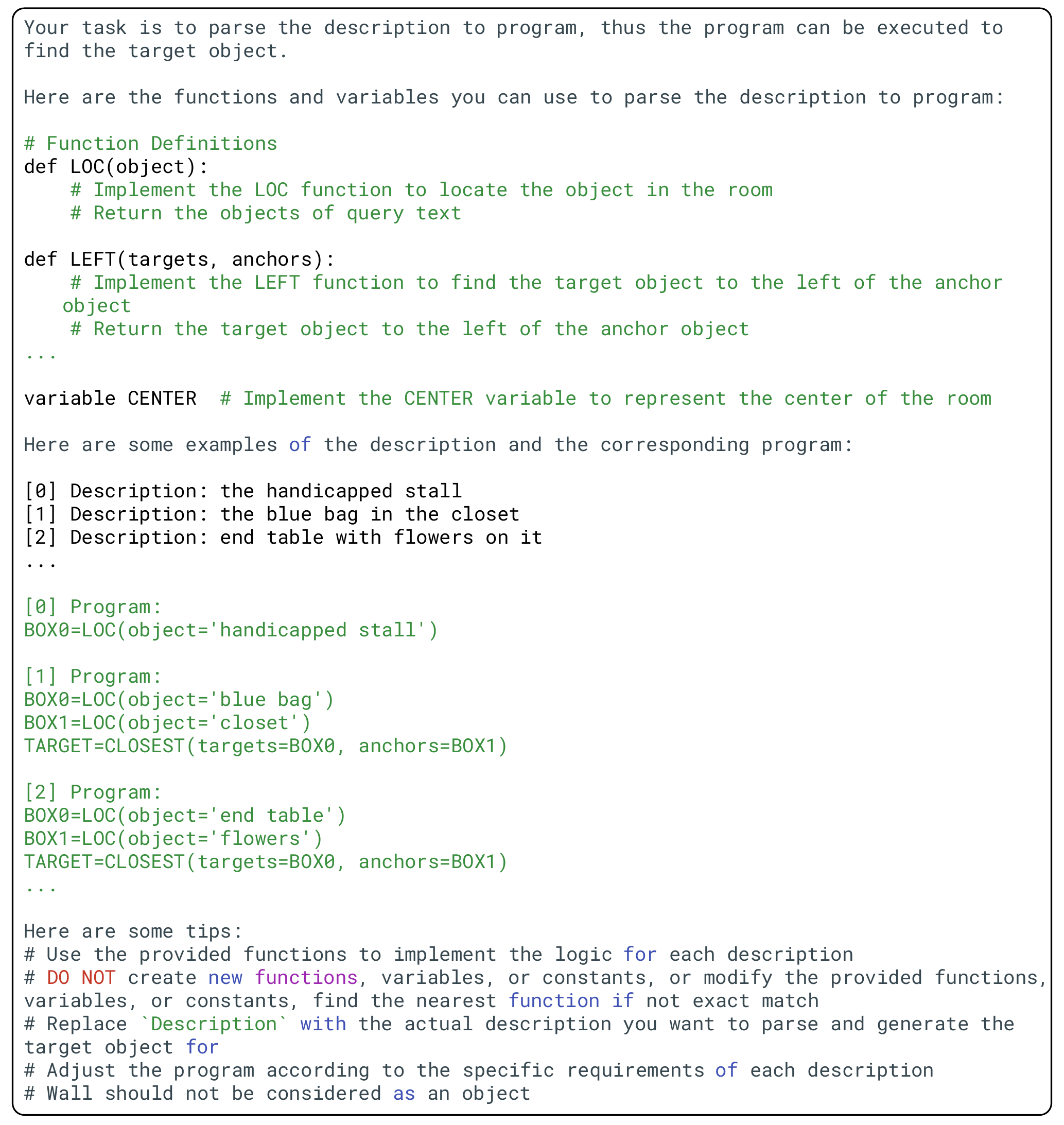}
    \caption{Prompt for generating visual programs.}
    \label{fig:supp_4}
\end{figure*}
\vspace{-10pt}
\section{Visualization Examples for Zero-shot 3DVG}
\label{sec:1}
We provide three examples to visualize the effectiveness of the proposed two zero-shot 3DVG approaches, i.e., dialog with LLM and visual programming.
Concretely, the first example, i.e., Figure ~\ref{fig:supp_1}, confirms that LLMs can effectively perform zero-shot 3DVG while also delivering commendable results. The second example, as illustrated in Figure~\ref{fig:supp_2}, shows that LLMs may encounter limitations in the tasks requiring spatial reasoning.
However, this issue can be effectively addressed by the visual programming approach. 
The third example, i.e., Figure~\ref{fig:supp_3}, further exemplifies that the visual programming approach is capable of executing multi-step reasoning, which involves initially identifying blinds that are positioned above the monitors, followed by selecting the desired one among them.

\section{Prompts for Generating 3D Visual Program}\label{sec:2}
As illustrated in Figure~\ref{fig:supp_4}, the prompts for generating 3D visual programs include four components as follows:
\begin{itemize}
    \item \textit{Task explanation}: We first describe the 3DVG task in natural language and provide it to the LLMs.
    \item \textit{Function and variable definition}: We define a set of functions and variables corresponding to the modules in the visual programming approach, such as LOC and LEFT.
    \item \textit{In-context examples}: We provide contextual examples illustrating how visual programs are structured and applied to guide LLMs.
    \item \textit{Best practices and tips}: We conclude with essential tips and best practices to ensure the effectiveness of the programs, highlighting the key aspects that guarantee optimal performance.
\end{itemize}

These four components collaboratively facilitate the LLM to understand the task requirement, thereby allowing it to construct effective visual programs for the 3DVG task.

\end{document}